\documentclass[
  a4paper,
  10pt,
  twoside,
  titlepage
]{scrartcl}

\usepackage[utf8]{inputenc}
\usepackage[T1]{fontenc}
\usepackage{XCharter}

\usepackage{xcolor}
\usepackage{graphicx}
\usepackage{amsmath}

\usepackage{authblk}

\usepackage[hidelinks]{hyperref}
\hypersetup{
  pdftitle={Energy-Efficient Information Representation in MNIST Classification Using Biologically Inspired Learning},
  pdfauthor={Patrick Stricker, Florian Röhrbein, Andreas Knoblauch},
  pdfkeywords={Training, Neuroplasticity, Bridges, Competitive learning, Perturbation methods, Biological system modeling, Neurons}
}

\begin{document}

\begin{titlepage}
\thispagestyle{plain}
\raggedright
\vspace*{.05\textheight}

{\Large\bfseries
Energy-Efficient Information Representation in MNIST Classification Using Biologically Inspired Learning
\par}

\vspace{\baselineskip}
{\large
Freiburg September 2024
\par}

\vspace{\baselineskip}

{\normalsize
Patrick Stricker\textsuperscript{*\dag}, Florian Röhrbein\textsuperscript{\dag}, Andreas Knoblauch\textsuperscript{*}
\par}

\vspace{.5\baselineskip}
{\small
\textsuperscript{*}KEIM Institute, Albstadt-Sigmaringen University, Germany\\
\textsuperscript{\dag}Department of Computer Science, Chemnitz University of Technology, Germany
\par}

\end{titlepage}

\section*{Abstract}

Efficient representation learning is essential for optimal information storage and classification. However, it is frequently overlooked in artificial neural networks (ANNs). This neglect results in networks that can become overparameterized by factors of up to 13, increasing redundancy and energy consumption. As the demand for large language models (LLMs) and their scale increase, these issues are further highlighted, raising significant ethical and environmental concerns. We analyze our previously developed biologically inspired learning rule using information-theoretic concepts, evaluating its efficiency on the MNIST classification task. The proposed rule, which emulates the brain's structural plasticity, naturally prevents overparameterization by optimizing synaptic usage and retaining only the essential number of synapses. Furthermore, it outperforms backpropagation (BP) in terms of efficiency and storage capacity. Furthermore, it eliminates the need for pre-optimization of network architecture, enhances adaptability, and reflects the brain's ability to reserve 'space' for new memories. This approach not only advances scalable and energy-efficient AI but also provides a promising framework for developing brain-inspired models that optimize resource allocation and adaptability.

\section{Introduction}

Deep neural networks (DNNs) have revolutionized fields such as computer vision, natural language processing, and speech recognition with their impressive performance in supervised learning tasks \cite{lecun2023, alam2020, he2015, lecun2015}. Nevertheless, DNNs frequently encounter difficulties associated with overparameterization, largely due to the use of backpropagation (BP). The intrinsic characteristics of gradient-based optimization inevitably lead to the storage of redundant information and potential noise, assigning nonzero weights to each synapse regardless of network size. Although DNNs are inspired by the human brain, current models are unable to fully replicate the brain's efficient generalization and learning capabilities by focusing on the synaptic weight plasticity as the sole mechanism \cite{ijcnn}. In contrast, the human brain demonstrates efficient information processing through sparse neural connections, driven by both spatial and energy constraints \cite{knoblauch, tiddia}. Sparse connectivity enhances generalization by reducing the storage of noise and redundant information. Based on Alemi et al.'s \cite{alemi} discovery that compression correlates with improved generalization, we suggest that this might explain the brain's superior generalization ability.

While methods such as dropout layers and skip connections can mitigate some effects of overparameterization, it should be noted that these techniques serve to mask the drawbacks of increased model size. Han et al. \cite{han2015} demonstrated, that a significant proportion of large architectures contain considerable redundancy, with sizes far exceeding what is necessary for optimal performance. This overparameterization results in inflated computational resource demands and increased carbon emissions, exacerbating environmental and ethical concerns, particularly given the rapid growth of generative AI. Recent studies on models like Llama and Minitron \cite{nvidia} have highlighted the severity of this issue, showing that pruned Large Language Models (LLMs) can achieve comparable or superior performance while using 40-150 times fewer training tokens. The short lifespan of many LLM deployments, combined with their substantial resource consumption, intensifies the need for more efficient and sustainable approaches. Thus, there is a critical need for algorithms that deliver high performance without unnecessary complexity. Our proposed framework addresses this by focusing on efficient information representation, emulating the brain's ability to achieve efficient performance with sparse connectivity. This approach reduces model size and resource usage, providing a solution to the issues of overparameterization and excessive resource consumption.

In our previous work \cite{ijcnn}, we introduced a biologically inspired learning framework that not only achieves performance comparable to that of BP but also offers substantial benefits in energy-efficient information representation. The framework retains only those synapses, that are essential for the classification task, thereby emulating the efficiency observed in the brain. The selective retention of synapses, a process known as structural plasticity \cite{knoblauch, tiddia, larosa2020, janzakova2023}, reflects the brain's capacity to optimally and efficiently store information. This contrasts with BP, which scales by utilising all available synapses.

This paper extends our earlier research \cite{ijcnn} by focusing specifically on the efficiency of information representation in the MNIST classification task. We frame our learning framework as a neural associative network, which frequently serves as a computational model of brain functions \cite{memory1}. Using this framework, we reinterpret the MNIST classification task as a heteroassociative memory problem, where patterns are linked and retrieved through sparse synaptic connections. This approach enables the quantification of information storage and compression. Our approach is predicated on the assumption that the neural network forms a Markov chain, following the insights of Tishby and Zaslavsky \cite{tishby2015, tishby2017}, and integrates a Variational Information Bottleneck (VIB) layer \cite{b22}. The stochastic encoding layer learns the distribution of hidden layer activations, pretrained with our approach or the benchmark. This allows for a more precise calculation of mutual information, allowing for a deeper analysis of information representation.

The results are benchmarked against a range of methods, including Chorowski's \cite{b1} approach, which applies nonnegativity and sparsity constraints to BP, as well as standard BP. While Chorowski's methods demonstrate superior accuracy, our approach is shown to outperform both in terms of storage efficiency and classification performance relative to the number of nonsilent synapses used.

The application of biologically inspired learning rules has the potential to lead to the development of more sustainable models. Thus, drawing inspiration from these biologically inspired rules could assist in addressing the growing need for energy efficiency in AI. We provide evidence that our framework maintains comparable classification performance, significantly improves storage efficiency, and reduces network size. 

\section{Modeling}
\subsection{The Memory Task}
Memories are commonly identified with patterns of neural activity that can be revisited, evoked, or stabilized by appropriately modified synaptic connections \cite{knoblauch}. In the simplest case, such a memory corresponds to a group of neurons that fire at the same time and, according to the Hebbian hypothesis that 'what fires together wires together' \cite{hebb}, develop strong mutual synaptic connections \cite{knoblauch, caporale, clopath, knoblauch2012}. These groups of strongly connected neurons, known as cell assemblies \cite{knoblauch, hebb, palm2014}, are closely related to associative networks and play an important role in neuroscience as models of neural computation for various brain structures.

Formally, networks of cell assemblies can be modeled as associative networks, which are essentially single-layer neural networks employing Hebbian learning \cite{knoblauch}. The concept, first introduced by Steinbuch in 1961, featured a single-layer network where neurons receive inputs from an address pattern and establish associations directly using binary neurons and synapses \cite{steinbuch1961}. The model can be extended to a two-layer architecture, thereby enabling the processing of more complex patterns. In this setup, an address pattern \( u \) is mapped to a hidden layer that extracts intermediate representations, which are then projected to a content pattern \( v \). The synaptic matrix \( A \) is updated locally according to Hebbian learning principles.

In our study, MNIST classification is framed as a heteroassociative memory task using a two-layer network. The first layer captures intermediate representations of image patterns \( u \), while the second layer maps these representations to the final labels \( v \). Heteroassociation in this context involves learning and storing classification patterns in synaptic weight matrices, which serve as the memory matrix \( A \), with negative weights clipped to zero to ensure nonnegativity \cite{memory1, kohonen1977}. This method effectively emulates associative memory processes, reflecting how neural networks perform pattern recognition and information retrieval \cite{memory1}.

Recent research utilizing FashionMNIST has demonstrated the effectiveness of heteroassociative learning rules for pattern denoising, showing how associative memory principles can be applied in classification tasks \cite{kymn2024}. The study revealed that models leveraging associative memory could effectively denoise structured patterns, emphasizing the relevance of such methods for classification tasks. This research supports the applicability of MNIST datasets for similar tasks, as MNIST and FashionMNIST exhibit comparable structural and complexity characteristics.

\subsection{Competitive Hebbian Plasticity and Weight Perturbation}\label{hebbian}

To address the heteroassociative memory problem in MNIST classification, we employ our previously developed biologically inspired learning framework \cite{ijcnn}. This framework integrates multiple plasticity rules, including competitive excitatory Hebbian plasticity \cite{b1}, nonnegativity constraints, and weight perturbation (WP) \cite{ijcnn, cauwenberghs, DemboKailath, test1, saran}. These mechanisms ensure that learning is both efficient and biologically plausible, promoting sparsity in hidden layers and maintaining nonnegativity across synaptic weights. The classification layer further employs homeostatic plasticity through bias neurons, where the bias values act as thresholds to enhance class discrimination while ensuring stable convergence \cite{ijcnn, b1}.

The modification of synaptic weights $w_{ij}$ between a neuron $i$ and a neuron $j$ in the layer above in the hidden layer is given by:
\begin{equation}
	\Delta w_{ij}=\eta z_j\cdot(x_i - \sum_{k\neq j} z_kw_{ik}),
\label{local}
\end{equation}
where $x_i$ represents the input from neuron $i$, $\eta$ denotes the learning rate, and $z_j = \sum_{u=1} x_uw_{uj}$ refers to the resulting activity of neuron $j$.

The classification layer is updated via the following equation:
	\begin{equation}
	\Delta w_{ki} = \eta \alpha\cdot\Delta w^{hebbian}_{ki}+\eta\beta\cdot\Delta w^{WP}_{ki},
	\label{classi}
	\end{equation}
where the parameters $\alpha \text{ and } \beta$ denote the contribution of the corresponding plasticity mechanism. The WP component, \( \Delta w_{ki}^{WP} \), is calculated using the following update rule:
\begin{equation}
\Delta w_{ij}^{\text{WP}} = - \frac{\eta}{\sigma^2} (E^{pert} - E) \xi_{ij},
\label{eq:8}
\end{equation}
where \( E^{pert} \) refers to the error of the perturbed trial, \( E \) denotes the error of the unperturbed trial, \( \xi_{ij} \) represents the perturbation term, and \( \sigma^2 \) reflects the strength of the perturbation. In this approach, weights are adjusted based on performance changes from perturbations, where improvements lead to updates in the direction of perturbations, and degradations result in opposite adjustments \cite{test1}.

The bias weights are updated according to the following equation:
	\begin{equation}
	\Delta{b_{k}} = \eta \gamma\left(\frac{1}{K}\cdot\sum_{k=1}^Kz_{kt}-\frac{1}{N}\cdot\sum_{t=1}^Nz_{kt}\right),
	\label{synscaw}
	\end{equation}
where the first term denotes the mean activation of the layer, the second term represents the mean activation of the current minibatch, and $\gamma$ controls the contribution of this learning rule\cite{b22}. As this work focuses on classification tasks, training batches with equally distributed labels are assumed. This simplifies the mean target activation to $\frac{1}{K}$\cite{ijcnn}.

\subsection{Information Theory for Deep Associative Neural Networks}\label{encoder}

To quantify the mutual information within a multilayer network, we adopt an information-theoretic perspective inspired by Tishby and Zaslavsky \cite{tishby2015, tishby2017}. They show that neural networks form a Markov chain of successive representations, where each layer \( T \) can be treated as a random variable with its encoder \( P(T \mid X) \) and decoder \( P(Y \mid T) \). This clarifies the flow of information across network layers and aligns inherently with our approach. The Markov chain model's assumption that each layer's state depends only on the immediately preceding layer is naturally satisfied by our Hebbian learning rule, which updates weights based on local activity

Traditional methods for estimating mutual information, such as bin discretization, can be biased and sensitive to bin size \cite{b22, poole2019}. To address these issues, we opted to obtain variational approximations of mutual information using a Variational Autoencoder (VAE)-based approach \cite{b22}.

In this study, we depart from the conventional methodology by incorporating a deterministic bottleneck in the form of a pretrained hidden layer preceding the variational encoder, while setting the $\beta$ parameter to zero. This adjustment removes the typical information bottleneck, ensuring that compression is solely induced by the frozen hidden representations while still allowing for accurate mutual information estimation.

In order to calculate and compare the information representation and compression induced by different learning rules, it is first necessary to extract a deterministic hidden representation, designated as \( H \), from a neural network which has been pretrained using either our approach, BP, or constrained BP. We then use a stochastic encoding layer to model \( p(Z \mid H) \), where \( Z \) is the latent variable. This encoding layer is parameterized by \( 2K \) parameters, where \( K \) is the size of the layer. Specifically, the layer outputs means \( \mu \) and variances \( \sigma \) (after applying a softplus transformation) for \( Z \), approximating its distribution as \( \mathcal{N}(Z \mid \mu, \sigma) \) \cite{alemi}. This setup ensures that the encoder effectively captures the distribution of \( Z \). We then estimate the mutual information \( I(X; Z) \) using the Kullback–Leibler divergence \( D_{KL}[p(Z \mid X) \| p(Z)] \), following the approach described in \cite{alemi, tishby2015, tishby2017, b22}.

\subsection{Performance Measures for Associative Memory}

Biological neural networks achieve efficient memory storage through sparse connectivity, which is driven by both spatial and energy constraints. The energy cost associated with neural signaling is closely linked to the maintenance of nonsilent synapses, which are crucial for memory retention \cite{knoblauch, lennie2003, laughlin2003, attwell2001}. Inspired by these principles, we evaluate neural network performance in terms of associative memory capabilities using the synaptic capacity \( C^S \) metric introduced by Knoblauch et al. \cite{knoblauch, memory1}. This metric quantifies memory efficiency by normalizing channel capacity against the number of nonsilent synapses. Synaptic capacity \( C^S \) is defined as:

\begin{equation}
C^S = \frac{I(Z; X)}{\text{Number of nonsilent synapses}} \text{ [bits/synapse]},
\label{capa}
\end{equation}

where \( I(Z; X) \) represents the mutual information between the latent variable \( Z \) and the input \( X \). This measure reflects how effectively the network uses its connections for memory storage and retrieval while minimizing the number of nonsilent synapses. By adopting \( C^S \), we align with the brain's approach of compressing information and reducing parameter count, thus providing insights into memory efficiency and network performance relative to synaptic resources.

\section{Numerical Experiments and Results}
\subsection{Experimental Setup}

This paper explores associative neural networks for MNIST classification, focusing on information representation through a heteroassociative memory approach. We employ a one-hidden-layer feedforward network where the hidden layer's weight matrix is updated using competitive Hebbian plasticity, as detailed in \autoref{hebbian}. The update rule for the hidden layer is given by \eqref{local}, while the classification layer is trained with a combination of \eqref{local} and \eqref{eq:8}, resulting in \eqref{classi}. Bias weights are updated according to \eqref{synscaw}. Experiments were conducted using the Keras framework with TensorFlow as the backend.

A modified sigmoid activation function is used in the hidden layer, transforming outputs from \([0.5, 1.0]\) to \([0.0, 1.0]\) to align with the nonnegative values processed by the network. We use batch processing to extend trials over time with duration \(T\), where each mini-batch consists of \(N_{batch}\) inputs indexed by \(t = 1,..., N_{batch}\). 

For our experiments, we use a subset of the MNIST dataset\footnote{The MNIST dataset comprises 70,000 grayscale images of handwritten digits (28 × 28 pixels), including 60,000 training and 10,000 test images \cite{mnist, b22}. Our focus is on the digits 1, 2, and 6.} and apply Min-Max scaling to standardize pixel values between 0 and 1. Weight matrices are initialized with values drawn from a random uniform distribution between 0.01 and 0.1, and the cross-entropy loss function is employed for optimization \cite{ijcnn}.

To assess storage and energy efficiency, we train networks using our approach, BP, or constrained BP. After training, we extract a deterministic hidden representation \( H \) from these networks and apply a stochastic encoding layer to model \( p(Z \mid H) \). We set \( K \) for each network architecture according to \( 2K = M \), where \( M \) is the number of hidden units. Mutual information \( I(X; Z) \) is then estimated from this stochastic encoding to evaluate information compression and synaptic capacity. We compare these results with benchmarks and analyze synaptic capacity to identify optimal network architectures for both storage and energy efficiency.

\subsection{Efficient Simulation Using BWHPC and Local Prototyping}
\begin{table}[t]
\caption{Mean epoch times and standard deviations for different GPUs (in seconds per epoch), measured during benchmarking of smaller-scale experiments.}
\label{performance_table}
\centering
\begin{tabular}{|c|c|c|c|c|}
\hline
\textbf{GPU} & \textbf{RTX 4080} & \textbf{H100} & \textbf{Tesla P100} & \textbf{A100} \\
\hline
\textbf{Time} & $2.255 \pm 0.114$ & $2.309 \pm 0.0476$ & $2.891 \pm 0.235$ & $2.307 \pm 0.073$ \\
\hline
\end{tabular}
\end{table}

Initial prototyping on a local NVIDIA RTX 4080 facilitated rapid development of smaller models with reduced batch sizes. However, larger batch sizes are crucial for our learning rule's effectiveness \cite{ijcnn}. Final simulations were conducted on the bwUniCluster 2.0, where GPUs like the A100, H100, and Tesla P100 provided the necessary scalability and performance for handling larger models and batch sizes that exceeded the RTX 4080's memory capacity. As shown in Table \ref{performance_table}, the A100 and H100 had comparable performance to the RTX 4080 on smaller models but offer much better scalability with larger batch sizes. All experiments used the Keras and TensorFlow frameworks to ensure consistency across different environments.

\subsection{Analysis of Network Storage Capacity and Connectivity Through Mutual Information}

\begin{table}[t]
    \caption{Performance of Different Algorithms with Various Hidden Units}
    \label{performance_table2}
    \centering
    \renewcommand{\arraystretch}{1.0} 
    \begin{tabular}{|c|l|r|r|r|}
    \hline
    \textbf{Hidden Neurons} & \textbf{Algorithm} & \textbf{Test Accuracy} & \textbf{I(X,Z)} & $\mathbf{C^S}$ \\
    \hline
    10 & BP   & \textbf{99.01\%} & 22.50 \text{ bits} & $2.87 \times 10^{-3}$ \\
       & Chorowski \cite{b1} & 98.34\% & 16.10 bits & $1.08 \times 10^{-2}$ \\
       & Authors   & 64.29\% & \textbf{12.80 bits} & $\mathbf{1.63 \times 10^{-2}}$ \\
    \hline
    30 & BP   & \textbf{99.17\%} & 127.62 \text{ bits} & $5.43 \times 10^{-3}$ \\
       & Chorowski \cite{b1} & 99.01\% & 72.99 \text{ bits}  & $2.22 \times 10^{-2}$ \\
       & Authors   & 86.21\% & \textbf{52.18 bits}  & $\mathbf{6.66 \times 10^{-2}}$ \\
    \hline
    100 & BP  & \textbf{99.23\%} & 435.93 \text{ bits} & $5.56 \times 10^{-3}$ \\
        & Chorowski \cite{b1} & 99.10\% & 217.43 \text{ bits} & $5.93 \times 10^{-2}$ \\
        & Authors   & 89.79\% & \textbf{198.33 bits} & $\mathbf{2.53 \times 10^{-1}}$ \\
    \hline
    200 & BP  & \textbf{99.17\%} & 518.79 \text{ bits} & $3.31 \times 10^{-3}$ \\
        & Chorowski \cite{b1} & 99.10\% & 402.91 \text{ bits} & $1.31 \times 10^{-1}$ \\
        & Authors   & 95.55\% & \textbf{372.66 bits} & $\mathbf{4.75 \times 10^{-1}}$ \\
    \hline
    \end{tabular}
    \caption*{Performance metrics for our algorithm compared to two benchmark algorithms. Metrics include accuracy (as a percentage), mutual information \(I(X, Z)\) in bits, and synaptic capacity in bits per nonsilent synapse. The best values in each category are highlighted in bold. Benchmarks were replicated to ensure comparability. Hyperparameters for our algorithm are: \(\eta = 0.000158\), \(\alpha = 0.1\), \(\beta = 446.25\), \(\gamma = 0.1\), \(\sigma^2 = 0.0157\).}
\end{table}

Table~\ref{performance_table2} summarizes the results for various network topologies and training algorithms. As shown in our previous work \cite{ijcnn}, increasing the number of hidden neurons \(M\) nearly eliminates the performance disparity between our learning rule and constrained BP. The remaining discrepancy is likely due to the precise gradient calculations and negative biases used by BP, which are not present in our framework.

Notably, the improvement observed in the baseline scenario compared to our prior work is attributed to the use of a modified sigmoid activation function. This adjustment led to significant improvements in performance, addressing issues where training a model with the baseline was previously infeasible \footnote{For a detailed analysis of performance metrics and further insights, please refer to \cite{ijcnn}.}.

The mutual information \(I(X, Z)\) increases with layer size, consistent with our approach that employs stochastic encoding correlated with hidden layer size. Our method achieved the most effective information compression in all cases. While Chorowski et al.'s \cite{b1} technique demonstrates superior compression compared to BP in the smallest architecture due to its high degree of sparsification, our method consistently outperforms both Chorowski et al.'s and BP across all tested architectures. BP retains the most information overall by keeping all synaptic weights nonzero, which can lead to the storage of substantial amounts of redundant information and noise, given its ability to perfectly memorize any dataset with sufficient parameters and training epochs. These findings align with Alemi et al. \cite{alemi}, who observed that increased mutual information \(I(X, Z)\) often correlates with reduced generalization.

For sparse algorithms, such as constrained BP and our approach, synaptic capacity \(C^S\) increases with layer size due to enhanced storage in the stochastic encoding. BP, with all synapses being nonzero, exhibits the lowest synaptic capacity, indicating overparameterization and the retention of redundant and noisy information. In contrast, our algorithm achieves the highest synaptic capacity across all settings, efficiently retaining only essential synapses while storing fewer bits compared to other methods. This demonstrates that our method results in the most efficient information representation, with the fewest nonzero synapses and lowest stored information across all scenarios.

\section{Discussion}

We evaluated the effectiveness of our biologically inspired learning rule for associative neural networks by analyzing its impact on information representation and synaptic capacity across various network topologies and training algorithms. Our findings indicate that our method significantly improves synaptic capacity compared to both constrained and unconstrained BP, especially in larger architectures. Our approach maintains high classification accuracy while efficiently utilizing synaptic resources. Additionally, our method's ability to balance compression with synaptic capacity results in superior energy efficiency compared to methods that focus solely on minimizing mutual information.

In contrast to biologically implausible compression algorithms like constrained BP and VAE, which focus solely on minimizing mutual information, our approach adopts a dual strategy to maximize both compression and synaptic capacity. Inspired by neuroplasticity in the human brain, which balances spatial and energy constraints for efficient information processing \cite{knoblauch}, our method prevents overparameterization while enhancing both efficiency and performance.

Our biologically inspired learning rule effectively models brain mechanics by optimizing synaptic capacity and achieving energy-efficient information representation. In contrast, methods like constrained BP often retain redundant information as network size exceeds the minimal number of hidden neurons necessary for satisfactory classification accuracy. Despite using nonnegativity constraints and activity regularization, these methods struggle with overparameterization. Our approach overcomes these limitations by using local competition and nonnegativity to ensure that weights converge to solutions primarily determined by the input data's covariance matrix \cite{b2}, thereby effectively learning the minimal sufficient statistics.

While traditional theories attribute adult learning and memory to Hebbian modification of synaptic weights, recent research suggests that structural plasticity, involving network rewiring and the generation or removal of synapses, also plays a crucial role in learning and memory \cite{knoblauch, tiddia, hebb, bliss1993, paulsen2000, song2000, navlakha2015, zito2002, knoblauch2017}. In order to model these processes, researchers have developed mathematical frameworks that simulate network dynamics, to evaluate the impact of such structural changes on memory and performance. Knoblauch et al. \cite{knoblauch} employ a Markov approach to define synapses in three states: potential, instantiated but silent, and instantiated and stabilized. Structural plasticity in their model relates to transitions between these states \cite{knoblauch, tiddia}. Tiddia et al.'s \cite{tiddia} approach structural plasticity through synaptic rewiring, where synapses below a predefined threshold are pruned and rewired at regular intervals.

Although these methodologies are based on biological principles, they do not fully capture the brain's adaptability. Once synapses are consolidated or stabilized in these models, they cannot be pruned or adjusted further. In contrast, the brain continuously rewires to accommodate new memories while maintaining synaptic density \cite{navlakha2015, zito2002}. Additionally, these methods require explicit modeling of synaptic retention rates and the optimization of network architecture before achieving optimal information representation.

In contrast, our method inherently promotes these properties through its plasticity mechanisms, integrating compression and capacity optimization without the need for additional constraints. By dynamically regulating retained parameters based on data and signal dynamics, our approach maintains efficiency while reflecting the brain's ability to reserve 'space' for new memories and preserve adaptability even in a stable state. This results in a more flexible and energy-efficient model, offering insights into the brain’s resource allocation strategies. While our method achieves slightly inferior classification performance compared to BP, it closely mirrors the brain’s adaptive learning processes. Thus, it presents a promising approach for emulating efficient associative memory and advancing our understanding of neural resource management. We are actively researching methods to close this performance gap and enhance the overall efficacy of our approach.

\section{Conclusion}

Our study investigates a biologically inspired learning rule that utilizes Competitive Hebbian plasticity principles to enhance energy-efficient information representation in DNNs. By optimizing synaptic capacity and managing redundancy, our approach mirrors the brain's strategy of maintaining sparse and effective neural connections. In contrast to traditional methods that struggle with overparameterization and redundant information, our method achieves significant advancements in storage efficiency, network size reduction, and energy consumption. This approach demonstrates the potential of biologically inspired learning rules to develop more sustainable and scalable AI models, meeting the growing demand for energy-efficient and ethical AI solutions.

Although our method exhibits slightly inferior classification performance compared to BP, it provides a promising approach for emulating efficient associative memory and advancing our understanding of neural resource management. By dynamically regulating retained parameters based on data and signal dynamics, our approach offers a potential explanation for how the brain optimally allocates neurons and synapses while avoiding overparameterization. This method also reserves capacity for future learning, reflecting the brain's adaptive processes. We are actively working to close the performance gap and further improve our approach.

Future research will address the performance limitations identified in this study by enhancing classification accuracy and computational efficiency, while maintaining the benefits of our approach. We will also investigate the model's ability to learn multiple tasks within a single network, emulate broader brain generalization capabilities, and assess its effectiveness on more complex architectures and deeper DNNs to evaluate scalability and performance under increased network complexity.

\end{document}